\documentclass[12pt,stdletter,orderfromtodate,sigleft,dateno]{article}

\usepackage{arxiv}

\usepackage[utf8]{inputenc} 
\usepackage[T1]{fontenc}    
\usepackage{hyperref}       
\usepackage{url}            
\usepackage{booktabs}       
\usepackage{amsfonts}       
\usepackage{amsmath}
\newcommand\numberthis{\addtocounter{equation}{1}\tag{\theequation}}
\usepackage{nicefrac}       
\usepackage{microtype}      
\usepackage{lipsum}
\usepackage{graphicx}
\usepackage[natbibapa]{apacite}
\graphicspath{ {./images/} }

\title{Whence the Expected Free Energy?}

\author{
 Beren Millidge \\
  School of Informatics\\
  University of Edinburgh\\
  \texttt{beren@millidge.name}
   \And
 Alexander Tschantz \\
  Sackler Center for Consciousness Science\\
  School of Engineering and Informatics\\
  University   Sussex \\
  \texttt{tschantz.alec@gmail.com} \\
 \And
 Christopher L Buckley \\
  Evolutionary and Adaptive Systems Research Group\\
  School of Engineering and Informatics\\
  University of Sussex\\
  \texttt{C.L.Buckley@sussex.ac.uk} \\
}
\date{\vspace{-5ex}}

\begin{document}
\maketitle
\begin{abstract}
The Expected Free Energy (EFE) is a central quantity in the theory of active inference. It is the quantity that all active inference agents are mandated to minimize through action, and its decomposition into extrinsic and intrinsic value terms is key to the balance of exploration and exploitation that active inference agents evince. Despite its importance, the mathematical origins of this quantity and its relation to the Variational Free Energy (VFE) remain unclear. In this paper, we investigate the origins of the EFE in detail and show that it is not simply "the free energy in the future". We present a functional that we argue is the natural extension of the VFE, but which actively discourages exploratory behaviour, thus demonstrating that exploration does not directly follow from free energy minimization into the future. We then develop a novel objective, the Free-Energy of the Expected Future (FEEF), which possesses both the epistemic component of the EFE as well as an intuitive mathematical grounding as the divergence between predicted and desired futures.
\end{abstract}
\section{Introduction}

The Free-Energy Principle (FEP) \citep{friston2010free,friston2006free,friston2012free} is an emerging theory from theoretical neuroscience which offers a unifying explanation of the dynamics of self-organising systems \citep{parr2020markov,friston2019physics}. It proposes that such systems can be interpreted as embodying a process of variational inference which minimizes a single information-theoretic objective -- the Variational Free-Energy (VFE). In theoretical neuroscience, the FEP translates into an elegant account of brain function \citep{friston2005theory,friston2003learning,friston2008hierarchical,friston2008DEM,friston2008variational}, extending the Bayesian Brain hypothesis \citep{knill2004Bayesian,doya2007Bayesian,deneve2005Bayesian} by postulating that the neural dynamics of the brain perform variational inference. Under certain assumptions about the forms of the densities embodied by the agent, this theory can even be translated down to the level of neural circuits in the form of a biologically plausible neuronal process theory \citep{bastos2012canonical, friston2008hierarchical,kanai2015cerebral,shipp2016neural,spratling2008reconciling}.

Action is then subsumed into this formulation, under the name of \textit{active inference} \citep{friston2009reinforcement,friston2011optimal,friston2012free} by mandating that agents act so as to minimize the VFE with respect to action \citep{buckley2017free,friston2006free}. This casts action and perception as two aspects of the same imperative of free-energy minimization, resulting in a theoretical framework for control which applies  to a variety of continuous-time tasks \citep{friston2011action,baltieri2017active,baltieri2018probabilistic,calvo2017predicting,millidge2019implementing}. 

Recent work has extended these ideas to account for inference over temporally extended action sequences. \citep{friston2016active,friston2015active,friston2017active,friston2012free,tschantz2019learning}. Here it is assumed that rather than action minimising the instantaneous VFE, sequences of actions (or policies) minimise the cumulative sum over time of a quantity called the \textit{Expected Free Energy} (EFE) \citep{friston2015active}. Active inference using the EFE has been applied to a wide variety of tasks and applications, from modelling human and animal choice behaviour \citep{friston2015active,fitzgerald2015active,pezzulo2016active}, simulating visual saccades and other `epistemic foraging behaviour' \citep{mirza2016scene,parr2017active,parr2018active,friston2018deep,friston2017curiosity}, solving reinforcement learning benchmarks \citep{tschantz2019scaling,millidge2019deep,millidge2019combining,ccatal2020learning,van2019simulating,ueltzhoffer2018deep}, to modelling psychiatric disorders as cases of aberrant inference \citep{cullen2018active,mirza2019impulsivity,parr2018computational}. Like the continuous-time formulation, active inference also comes equipped with a biologically plausible process theory with variational update equations which have been argued to be homologous with observed neural firing patterns \citep{friston2017active,friston2017process,friston2017graphical,parr2019neuronal}.

A key property of the EFE is that it decomposes into both an extrinsic, value-seeking, and an intrinsic (epistemic), information-seeking term \citep{friston2015active}. The latter mandates active inference agents to resolve uncertainty by encouraging the exploration of unknown regions of the environment, a property which has been extensively investigated  \citep{friston2015active,schwartenbeck2013exploration,schwartenbeck2019computational,friston2017active}. The fact that intrinsic drives naturally emerge from this formulation is argued as an advantage over other formulations that typically encourage exploration by adding ad-hoc exploratory terms to their loss function \citep{oudeyer2009intrinsic,pathak2017curiosity,burda2018large,mohamed2015variational}. While the EFE is often described as a straightforward extension to the free energy principle that can account for prospective policies and is typically expressed in similar mathematical form  \citep{friston2015active,parr2017uncertainty,friston2017active,parr2019generalised,da2020active}, its origin remains obscure. Minimization of the EFE is sometimes motivated by a \textit{reductio ad absurdum} argument following from the FEP \citep{friston2015active,parr2019generalised} in that agents are driven to minimize the VFE, and therefore the only way they can act is to minimize their free-energy into the future. Since the future is uncertain, however, instead they must minimize the \textit{expected} free energy. Central to this logic is the formal identification of the VFE with the EFE. 

In this paper, we set out to investigate the origin of the EFE and its relations with the VFE. We provide a broader perspective on this question, showing that the EFE is not the only way to extend the VFE to account for action-conditioned futures. We derive an objective which we believe to be a more natural analogue of the VFE, which we call the \textit{Free Energy of the Future} (FEF), and make a detailed side-by-side comparison of the two functionals.Crucially, we show that the FEF actively discourages information-seeking behaviour, thus demonstrating that epistemic terms do not necessarily arise simply from extending the VFE into the future. We then investigate the origin of the epistemic term of the EFE, and show that the EFE is just the FEF minus the negative of the epistemic term in the EFE, which thus provides a straightforward perspective on the relation between the two functionals. We then propose our own mathematically principled starting point for action-selection under active inference -- the divergence between desired and expected futures, from which we obtain a novel functional the \textit{Free-Energy of the Expected Future} (FEEF), which has close relations to the generalized free energy \citep{parr2019generalised}. This functional has a natural interpretation in terms of the divergence between a veridicial and a biased generative model; it allows use of the same functional for both inference and policy selection, and it naturally decomposes into an extrinsic value term and an epistemic action term, thus maintaining the attractive exploratory properties of  EFE-based active inference while also possessing a mathematically principled starting point with an intuitive interpretation. 

\section{The Variational Free Energy}

The Variational Free Energy (VFE) is a core quantity in variational inference and constitutes a tractable bound on both the log model evidence and the KL divergence between prior and posterior \citep{beal2003variational,blei2017variational,fox2012tutorial,wainwright2008graphical}. For an in-depth motivation of the VFE and its use in variational inference, see Appendix \ref{variational}. 

The VFE, defined at time $t$, denoted by $\mathbf{F_t}$, is given by,
\begin{align*}
    \mathbf{F_t} & = \mathbf{D}_{KL}[Q(x_t | o_t; \phi)||p(o_t,x_t)] \\
    &= \mathbb{E}_{Q(x_t | o_t; \phi)} \big[ \ln \frac{Q(x_t | o_t; \phi)}{p(o_t,x_t)} \big] \numberthis \label{VFE_def}
\end{align*}

The agent receives observations $o_t$ and must infer the values of hidden states $x_t$. The agent assumes that the environment evolves according to a Markov process so that the distribution over states at the current time-step only depends on the state at the previous time-step, and that the observation generated at the current time-step depends only on the state at the current time-step. Given a distribution over a trajectory of states and observations, and under Markov assumptions it can be factorised as follows: $p(o_{0:T},x_{0:T}) = p(s_0) \prod_{t=0}^T p(o_t | s_t) p(s_{t+1} | s_t)$. In this paper, we also consider inference over future states and observations which have yet to be observed. Such future variables are denoted  $o_\tau$ or $x_\tau$ where $\tau > t$. To avoid dealing with infinite sums, agents only consider futures up to some finite time horizon, denoted $T$. $Q(x_t | o_t;\phi)$ denotes an approximate posterior density parametrised by $\phi$ which, during the course of variational inference, is fit as closely as possible to the true posterior. Note: there is a slight difference in notation here compared to that usually used in variational inference. Normally the approximate posterior is written as  $Q(x_t;\phi)$ without the dependence on $o$ made explicit. This is because the variational posterior is not a direct function of observations, but rather the result of an optimization process which depends on the observations.Here, we make the dependence on $o$ explicit to keep a clear distinction between the variational posterior $Q(x_t | o_t;\phi)$, obtained through optimization of the variational parameters $\phi$, and the variational prior $Q(x_t) = \mathbb{E}_{p(s_t | s_{t-1})}[Q(s_{t-1} | o_{t-1}; \phi)]$, obtained by mapping the previous posterior through the transition dynamics. Throughout this paper, we assume that inference is occurring in a discrete-time Partially-Observed Markov Decision Process (POMDP). This is to ensure compatibility with the EFE formulation later on, which is also situated within discrete-time POMDPs. \footnote{It is important to note that the original FEP was formulated in continuous time with generalised coordinates \citep{friston2006free,friston2008hierarchical} (where the hidden states are augmented with their temporal derivatives up to theoretically infinite order). The generalised coordinates mean that the agent is effectively performing variational inference over a Taylor-expanded future trajectory instead of a temporally-instant hidden state \citep{friston2008DEM,friston2008variational}. Action is derived by minimizing the gradients of the instantaneous VFE with respect to action, which requires the use of a forward model. More recent work on active inference and the FEP returns to the continuous-time formulation \citep{friston2019physics,parr2020markov} and the conclusions drawn in this paper may look different in the continuous-time domain.}   

The utility of the VFE for inference comes from the fact that the VFE is equal to the divergence between true and approximate posteriors up to a constant: $\mathbf{F}_t \geq \mathbf{D}_{KL}[Q(x_t | o_t; \phi)||p(x_t|o_t)]$. Thus, minimizing $\mathbf{F}_t$ with respect to the parameters of the variational distribution makes $Q(x_t;\phi)$ a good approximation of the true posterior.

One can also motivate the VFE as a technique to estimate model evidence. Log model evidence is a key quantity in Bayesian inference but is often intractable, meaning it cannot be computed directly. Intuitively, the log model evidence scores the likelihood of the data under a model, and thus provides a direct measure of the quality of a model. Under the free energy principle, minimizing the negative log model evidence (or \textit{surprisal}) is the ultimate goal of self-organising systems \citep{friston2006free,friston2012free,friston2012ao}. The VFE provides an upper bound on the log model evidence. This can be shown by importance sampling the  model evidence with respect to the approximate posterior, and applying Jensen's inequality:
\begin{align*}
     -\ln p(o_t) &= -\ln \int dx_t \, p(o_t,x_t) \\
    &=- \ln \int dx_t \, p(o_t,x_t) \frac{Q(x_t | o_t; \phi)}{Q(x_t | o_t; \phi)} \\
    &\leq -\int dx_t \, Q(x_t | o_t; \phi) \ln \frac{p(o_t,x_t)}{Q(x_t | o_t; \phi)} \\
    &\leq \mathbf{D}_{KL}[Q(x_t | o_t; \phi)||p(o_t,x_t)] \\
    &\leq  \mathbf{F_t}
\end{align*}
Since the VFE is an upper bound on the log model evidence (or surprisal), as the VFE is minimized, it becomes an increasingly accurate estimate of the surprisal.
To get a feel for the properties of the VFE, we showcase the following decomposition:
\begin{align*}
    \mathbb{F} & = \mathbf{D}_{KL}[Q(x_t | o_t; \phi)||p(o_t,x_t)]\\
    &= \mathbb{E}_{Q(x_t | o_t; \phi)} \big[ \ln  \frac{Q(x_t | o_t; \phi)}{p(o_t,x_t)}  \big] \\
    &= \underbrace{-\mathbb{E}_{Q(x_t | o_t;\phi)}[\ln p(o_t|x_t)]}_{\text{Accuracy}} + \underbrace{\mathbf{D}_{KL}[Q(x_t | o_t; \phi)||p(x_t)]}_{\text{Complexity}} \numberthis 
\end{align*}

This decomposition is the one typically used to compute the VFE in practice and has a straightforward interpretation. Specifically, minimizing the negative accuracy (and thus maximizing accuracy) ensures that the observations are as likely as possible under the states, $x_t$,  predicted by the variational posterior while simultaneously minimizing the complexity term, which is a KL divergence between the variational posterior and the prior. Thus the goal is to keep the posterior as close to the prior as possible while still maximizing accuracy. Effectively, the complexity term acts as an implicit regulariser, reducing the risk of overfitting to any specific observation.

\section{The Expected Free Energy}

While variational inference as presented above only allows us to perform inference at the current time given observations, it is possible to extend the formalism to allow for inference over actions or policies in the future.

To achieve this extension, a variational objective is required which can be minimized contingent upon future states and policies, which will allow the problem of adaptive action selection to be reformulated as a process of variational inference. To do this, the formalism must be extended in two ways. First, the generative model is augmented to include actions $a_\tau$, and policies, which are sequences of actions $\pi = [a_1, a_2...a_T]$. The action taken at the current time can affect future states, and thus future observations. In order to transform action selection into an inference problem, policies are treated as an inferred distribution $Q(\pi)$ which is optimised to meet the agents goals. The second extension required is to translate the notion of an agent's goals into this probabilistic framework. Active inference encodes an agent's goals as a desired distribution over observations $\tilde{p}(o_{\tau:T})$. We denote the biased distribution using a tilde over the probability density $\tilde{p}$ rather than the random variable to make clear that the random variables themselves are unchanged, it is only the agent's subjective distribution over the variables that is biased. \footnote{It is important to note that this encoding of preferences through a biased generative model is unique to active inference. Other variational control schemes \citep{rawlik2013stochastic,rawlik2013probabilistic,theodorou2010generalized,theodorou2012relative,levine2018reinforcement} instead encode desires through binary optimality variables and optimize the posterior given that the optimal path was taken. The relation between these frameworks is explored further in \citet{millidge2020relationship}.} This distribution is then incorporated into a biased generative model of the world $\tilde{p}(o_\tau,x_\tau) \approx \tilde{p}(o_\tau)Q(x_\tau | o_\tau)$ \footnote{Some more recent work \citep{friston2019physics,da2020active} prefers an alternative factorisation of the biased generative model in terms of an unbiased likelihood and a biased prior state distribution $\tilde{p}(o_\tau, x_\tau) = p(o_\tau | x_\tau)\tilde{p}(x_\tau)$. This leads to a different decomposition of the EFE in terms of risk and ambiguity (see Appendix \ref{decompositions}) but which is mathematically equivalent to the factorisation described here.}, where we have additionally made the assumption that the true posterior can be well approximated with the variational posterior: $p(x_\tau | o_\tau) \approx Q(x_\tau | o_\tau)$ which simply states that the variational inference procedure was successful \footnote{For additional information on the effect of this assumption, see appendix \ref{bound}.}.    Active inference proceeds by inferring a variational policy distribution $Q(\pi)$ that maximizes the evidence for this biased generative model. Intuitively, this approach turns the action selection problem on its head. Instead of saying: I have some goal, what do I have to do to achieve it? the active inference agent asks: Given that my goals were achieved, what would have been the most probable actions that I took?

A further complication of extending VFE into the future comes from the future observations. While agents have access to current observations (or data)  for planning problems, they must also reason about unknown future observations.  This is dealt with  by taking the expectation  of the objective with respect to  predicted observations $o_\tau$ drawn from the  generative model. 

In the active inference framework, the goal is to infer a variational distribution over both hidden states and policies that maximally fit to a biased generative model of the future. The framework defines the variational objective function to be minimized, the \textit{Expected Free Energy}, from time $\tau$ until the time horizon $T$, which is denoted $\mathcal{G}$:
\begin{align*}
    \mathcal{G} = \mathbb{E}_{Q(o_{\tau:T}, x_{\tau:T}, \pi)}[ \ln Q(x_{\tau:T}, \pi) - \ln \tilde{p}(o_{\tau:T},x_{\tau:T})]
\end{align*}

A temporal mean-field factorisation of the approximate posterior and of the generative model is assumed such that $Q(x_{\tau:T}, \pi) \approx Q(\pi) \prod_\tau^T Q(x_\tau)$ and $\tilde{p}(o_{\tau:T},x_{\tau:T}) \approx \prod_t^T \tilde{p}(o_\tau) Q(x_\tau | o_\tau)$. This factorisation neatly severs the temporal dependencies between time-steps. Given these assumptions, inferring the optimal $Q(\pi)$, turns out to be relatively straightforward.
\begin{align*}
     \mathcal{G} &= \mathbb{E}_{Q(o_{\tau:T}, x_{\tau:T}, \pi)} \big[ \ln Q(x_{\tau:T}, \pi) - \ln \tilde{p}(o_{\tau:T},x_{\tau:T}) \big] \\
     &= \mathbb{E}_{Q(o_{\tau:T},x_{\tau:T} | \pi)Q(\pi)} \big[ \ln Q(x_{\tau:T} | \pi) + \ln Q(\pi) - \ln \tilde{p}(o_{\tau:T},x_{\tau:T}) \big] \\
     &= \mathbb{E}_{Q(\pi)}[\ln Q(\pi) - \mathbb{E}_{Q(o_{\tau:T},x_{\tau:T} | \pi)} \big[ \sum_t^T [\ln Q(x_\tau) - \ln \tilde{p}(o_\tau, x_\tau)] \big] \\
     &= \mathbf{D}_{KL} \big[ Q(\pi) \Vert  e^{- \sum_t^T \mathcal{G}_\tau(\pi)} \big]
\end{align*}

Where  $\mathcal{G}_\tau(\pi) =  \mathbb{E}_{Q(o_\tau,x_\tau | \pi)}[\ln Q(x_\tau | \pi) - \ln \tilde{p}(o_\tau,x_\tau)]$ is defined to be the EFE for a single time--step $\tau$. From the KL-divergence above, it follows that the optimal variational policy distribution $Q^*(\pi)$ is simply the path integral into the future of the expected free energies for each individual time-step:
\begin{align*}
    Q^*(\pi) = \sigma(\sum_t^T \mathcal{G}_\tau(\pi)),
\end{align*}

where $\sigma(x)$ is a softmax function. This implies that to infer the optimal policy distribution it suffices to minimize the sum of expected free energies for each time step into the future. Inference proceeds by using the generative model to rollout predicted futures, computing the EFE of those futures, and then selecting policies which minimize the sum of the expected free energies. Since under temporal mean field assumptions, trajectories decompose into a sum of time-steps, it is sufficient for the rest of the paper to only consider a single time-step $\tau$.

To gain an intuition for the EFE, we showcase the following decomposition:
\begin{align*}
    \mathcal{G}_\tau(\pi) &=  \mathbb{E}_{Q(o_\tau,x_\tau | \pi)}[\ln Q(x_\tau | \pi) - \ln \tilde{p}(o_\tau,x_\tau)] \\
    &\approx \mathbb{E}_{Q(o_\tau,x_\tau | \pi)}[\ln Q(x_\tau | \pi) - \ln \tilde{p}(o_\tau) - \ln Q(x_\tau|o_\tau)] \\
    &\approx \underbrace{-\mathbb{E}_{Q(o_\tau,x_\tau |\pi)}\big[ \ln \tilde{p}(o_\tau) \big]}_{\text{Extrinsic Value}} -  \underbrace{\mathbb{E}_{Q(o_\tau)}\mathbf{D}_{KL}[Q(x_\tau | o_\tau)||Q(x_\tau | \pi)]}_{\text{Epistemic Value}} \numberthis  \label{EFE_decomp}
\end{align*}

While the EFE admits many decompositions, see Appendix \ref{decompositions} for a comprehensive overview,  the one presented in Equation 3 is perhaps the the most important because it separates the EFE into an extrinsic, goal-directed term (sometimes also called `instrumental value' in the literature) and an intrinsic, information-seeking term \footnote{The approximation in the final line of equation (3) is that we assume that the true and approximate posteriors are the same $Q(x_\tau | o_\tau) \approx p(x_\tau | o_\tau)$. Without this assumption, you obtain an additional KL divergence between the true and approximate posterior, which exactly quantifies the discrepancy between them (see appendix section 10 and 12 for more detail).}. The first term requires agents to maximize the likelihood of the desired observations $\tilde{p}(o_\tau)$ under beliefs about the future. It thus directs an agent to act to maximize the probability of its desires occurring in the future. It is called the extrinsic value term since it is the term in the EFE which accounts for the agent's preferences.

The second term in equation \ref{EFE_decomp} is the expected information gain, which is often termed the `epistemic value' since it quantifies the amount of information gained by visiting a specific state. Since the information gain is negative, minimizing the EFE as a whole mandates \textit{maximizing} the expected information gain.  This drives the agent to maximize the divergence between its posterior and prior beliefs, thus inducing the agent to take actions which maximally inform their beliefs and reduce uncertainty. It is the combination of extrinsic and intrinsic value terms which belies active inference's claim to have a principled approach to the exploration-exploitation dilemma \citep{friston2015active,friston2017active}.

The idea of maximizing expected information gain or "Bayesian surprise" \citep{itti2009bayesian} to drive exploratory behaviour has been argued for in neuroscience \citep{ostwald2012evidence,baldi2010bits}, and has been regularly proposed in reinforcement learning \citep{sun2011planning,still2012information,tschantz2020reinforcement,houthooft2016variational}. It is important to note however that in these prior works, information gain has often been proposed as an ad-hoc addition to an already existing objective function with only the intuitive justification of boosting exploration. In contrast, expected information gain falls naturally out of the EFE formalism, arguably lending the formalism a degree of theoretical elegance.

\section{Origins of the EFE}

Given the centrality of the EFE to the active inference framework, it is important to explore the origin and nature of this quantity. The EFE is typically motivated through a \emph{reductio ad absurdum} argument
 \citep{friston2015active,parr2019generalised}\footnote{An alternative motivation exists which situates the expected free energy in terms of a non-equilibrium steady state distribution \citep{da2020active,friston2019physics,parr2019computational}. This argument reframes everything in terms of a Gibbs free-energy, from which the EFE can be derived as a special case. The problem becomes, then, one of the motivation of the Gibbs free-energy as an objective function.}. The logic is as follows. Agents have prior beliefs over policies that drive action selection. By the FEP, all states of an organism, including those determining policies, must change so as to minimize free energy. Thus, the only self-consistent prior belief over policies is that the agent will minimize free-energy into the future through its policy selection process. If the agent did not have such a prior belief then it would select policies which did not minimize the free-energy into the future and would thus not be a free-energy minimizing agent. This logic requires a well-defined notion of the free-energy of future states and observations given a specific policy. The active inference literature implicitly assumes that the EFE is the natural functional which fits this notion \citep{friston2015active,friston2017process}. In the following section, we argue that the EFE is not in fact the only functional which can quantify the notion of the free energy of policy-conditioned futures, and indeed we propose a different functional \textit{The Free Energy of the Future}, which we argue is a more natural extension of the VFE to account for future states. 

\subsection{The Free Energy of the Future}

We argue that the natural extension of the free energy into the future must possess direct analogs to the two crucial properties of the VFE: it must be expressible as a KL-divergence between a posterior and a generative model, such that minimizing it causes the variational density to better approximate the true posterior. Secondly, it must also bound  the log model evidence of future observations. Bounding the log model evidence (or surprisal) is vital since the surprisal is the core quantity which, under the FEP, all systems are driven to minimize. If the VFE extended into the future failed to bound the surprisal, then minimizing this extension would not necessarily minimize surprisal, and thus any agent which minimized such an extension would be in violation of the FEP.  Here, we present a functional which we claim satisfies these desiderata -- the \textit{Free Energy of the Future} (FEF). 

We wish to derive an expression for variational free energy at some future
time $\tau$ that is conditioned on some policy $\pi$. In other words, we wish to quantify the free energy that will occur at some future time point, given some sequence of actions. Here, we derive a form of the ‘variational free energy of the future’, denoted $\mathbf{FEF}_\tau(\pi)$, by keeping the same terms as the VFE (Equation \ref{VFE_def}), but conditioning the variational distributions on our policy of interest and rewriting for the future time-point $\tau$.
Additionally, since observations in the future are unknown, we must evaluate our free energy under the expectation of our beliefs about future observations, as in the EFE. We thus define:

\begin{align*}
\mathbf{FEF}_\tau(\pi) =  \mathbb{E}_{Q(o_\tau, x_\tau | \pi)}[\ln Q(x_\tau | o_\tau) - \ln \tilde{p}(o_\tau,x_\tau)] 
\end{align*}
Since this equation is simply the KL-divergence between the variational posterior and the generative model, it satisfies the first desideratum. We next investigate the properties of the FEF by showcasing one key decomposition. As with the the VFE, we can then split the FEF into an energy and an entropy or an accuracy and complexity term, which correspond to the extrinsic and epistemic action terms in the EFE:
\begin{align*}
     \mathbf{FEF}_\tau(\pi) &= \mathbb{E}_{Q(o_\tau | \pi)}\mathbf{D}_{KL}[Q(x_\tau | o_\tau)||\tilde{p}(o_\tau,x_\tau)] \\
     &\approx -\underbrace{\mathbb{E}_{Q(o_\tau,x_\tau | \pi)} \big[ \ln \tilde{p}(o_\tau | x_\tau) \big]}_{\text{Accuracy}} + \underbrace{\mathbb{E}_{Q(o_\tau | \pi)}\mathbf{D}_{KL}[Q(x_\tau | o_\tau)||Q(x_\tau | \pi)]}_{\text{Complexity}}
\end{align*}
Unlike the EFE however, the expected information gain (complexity) term is positive while in the EFE term it is negative. Since the objective function, whether EFE or FEF, is to be \emph{minimized}, we see that using the FEF mandates us to minimize the information gain while the EFE requires us to maximize it (or minimize the negative information gain). An FEF agent thus tries to maximize its reward while trying to explore as little as possible. While this sounds surprising, it is in fact directly analogous to the complexity term in the VFE, which mandates maximizing the likelihood of an observation, while also keeping the posterior as close as possible to the prior \footnote{An objective functional equivalent to the FEF -- the "Predicted Free Energy" -- has also been proposed in \citep{schwobel2018active}. See section 14 of the appendix for more details.}.

\subsection{Bounds on the Expected Model Evidence}
We next show how the FEF can be derived as a bound on the expected model evidence satisfying the second desidaratum. We define the expected model evidence to be a straightforward extension of the model-evidence to unknown future states. The expected negative log model evidence for a trajectory from the current time-step $t$ to some time horizon $T$ is:
\begin{align*}
    -\mathbb{E}_{Q(o_{t:T} | \pi)}\big[ \ln \tilde{p}(o_{t:T}) \big]
\end{align*}

This objective states that we wish to maximize the probability (minimize the negative probability) of being in a desired trajectory $ \tilde{p}(o_{t:T})$, expected under the distribution of our beliefs about our likely future trajectories $Q(o_{t:T} | \pi)$ under a specific policy $\pi$.
Given a Markov generative model $p(o_{1:T},x_{1:T} | \pi) =\prod_t^T p(o_t | x_t)p(x_t | x_{t-1} | \pi)$, and assuming that the approximate posterior factorises $Q(x_{1:T}| o_{1:T}) = \prod_t^T Q(x_t | o_t)$, the expected model evidence factorises across time-steps,  it suffices to show the derivation for a single time-step $\tau > t$ (see Appendix \ref{trajectory} for a full trajectory derivation). We further define $Q(o_\tau, x_\tau | \pi) = Q(o_\tau | \pi)Q(x_\tau | o_\tau) = p(o_\tau | x_\tau)Q(x_\tau | \pi)$. We therefore take the expected model evidence for a single time-step, and show that the FEF is a bound on this quantity. 
\begin{align*}
    \label{FEF_bound}
    \numberthis
    - \mathbb{E}_{Q(o_\tau | \pi)}\big[ \ln \tilde{p}(o_\tau) \big] &= - \mathbb{E}_{Q(o_\tau | \pi)}\big[ \ln \int dx_\tau \tilde{p}(o_\tau,x_\tau) \big] \\
    &= - \mathbb{E}_{Q(o_\tau | \pi)}\big[ \ln \int dx_\tau \tilde{p}(o_\tau,x_\tau) \frac{Q(x_\tau | o_\tau)}{Q(x_\tau | o_\tau)} \big] \\
    &\leq - \mathbb{E}_{Q(o_\tau | \pi)}\int dx_\tau Q(x_\tau | o_\tau) \big[ \ln  \frac{\tilde{p}(o_\tau,x_\tau)}{Q(x_\tau | o_\tau)} \big] \\
    &\leq - \mathbb{E}_{Q(o_\tau,x_\tau | \pi)} \big[ \ln  \frac{\tilde{p}(o_\tau,x_\tau)}{Q(x_\tau | o_\tau)} \big] \\
    &\leq \mathbb{E}_{Q(o_\tau,x_\tau | \pi)} \big[ \ln  \frac{Q(x_\tau | o_\tau)}{\tilde{p}(o_\tau,x_\tau)} \big] \\
    &\leq \mathbb{E}_{Q(o_\tau | \pi)}\mathbf{D}_{KL}[Q(x | o_\tau)||\tilde{p}(o_\tau,x_\tau | \pi)] = \mathbf{FEF}(\pi)
\end{align*}

Crucially, this is an \textit{upper bound} on expected model evidence which can be tightened by minimizing the FEF. By contrast, returning to the EFE, we see below that since KL divergences are always $\geq 0$, the expected information gain is always positive, and so the EFE is a lower bound on the expected model evidence:
\begin{align*}
    \mathcal{G}_\tau(\pi) &=  \mathbb{E}_{Q(o_\tau,x_\tau | \pi)}[\ln Q(x_\tau | \pi) - \ln \tilde{p}(o_\tau,x_\tau)] \\
    &\approx \underbrace{-\mathbb{E}_{Q(o_\tau,x_\tau | \pi)}\big[ \ln \tilde{p}(o_\tau) \big]}_{\text{Negative Expected Log Model Evidence}} -  \underbrace{\mathbb{E}_{Q(o_\tau | \pi)}\mathbf{D}_{KL}[Q(x_\tau | o_\tau)||Q(x_\tau | \pi)]}_{\text{Expected Information Gain}}
\end{align*}

Since the expected information gain is an expected KL divergence, it must be $\geq 0$, and thus the negative expected information gain must be $\leq 0$. Since the EFE aims to minimize negative information gain (thus maximizing positive information gain), we can see minimizing the EFE actually drives it further from the expected model evidence. \footnote{There is a slight additional subtlety here involving the fact that there is also a posterior approximation error term which is positive. In general the EFE functions as an upper bound when the posterior error is greater than the information gain and a lower bound when the posterior error is smaller. Since the goal of variational inference is to minimize posterior error, and EFE agents are driven to maximize expected information gain, we expect this latter condition to occur rarely. For more detail see Appendix \ref{bound}.} 

We further investigate the EFE and its properties as a bound in Appendix \ref{bound}. Additionally, in Appendix \ref{naturalising} we review other attempts in the literature to derive the EFE as a bound on the expected model evidence and discuss their shortcomings.

\subsection{The EFE and the FEF}

To get a stronger intuition for the subtle differences between the EFE and the FEF, we present a detailed side-by-side comparison of the two functionals.
\begin{align*}
    \mathbf{FEF} &= \mathbb{E}_{Q(o_\tau,x_\tau | \pi)}[\ln Q(x_\tau | o_\tau) - \ln \tilde{p}(o_\tau,x_\tau)] \\
    \mathbf{EFE} &= \mathbb{E}_{Q(o_\tau,x_\tau | \pi)}[\ln Q(x_\tau | \pi) - \ln \tilde{p}(o_\tau,x_\tau)]
\end{align*}

While the two formulations might initially look very similar, the key difference is the variational term. The FEF, analogously to the VFE, measures the difference between a variational \textit{posterior} $Q(x_\tau | o_\tau)$ and the generative model $Q(x_\tau | \pi)$. The EFE, on the other hand, measures the difference between a variational \textit{prior} and the generative model. It is this difference which makes the EFE not a straightforward extension to the VFE for future time-steps, and underwrites its unique epistemic value term. 

We now demonstrate that both the EFE and the FEF can be decomposed into an expected likelihood, associated with extrinsic value, and an expected
KL-divergence between a variational posterior and a variational prior, associated with epistemic value. We factorise the generative model in the FEF into the (biased) likelihood and a variational prior, and factorise the generative model in the EFE into an approximate posterior, and a (biased) marginal:

\begin{align*}
    \mathbf{FEF} &= \mathbb{E}_{Q(o_\tau,x_\tau | \pi)}[\ln Q(x_\tau | o_\tau) - \ln \tilde{p}(o_\tau | x_\tau) - \ln Q(x_\tau | \pi)] \\
    \mathbf{EFE} &= \mathbb{E}_{Q(o_\tau,x_\tau | \pi)}[\ln Q(x_\tau | \pi) - \ln \tilde{p}(o_\tau) - \ln Q(x_\tau | o_\tau)]
\end{align*}

The variational prior and variational posterior can then be combined in both the FEF and the EFE to form epistemic terms. Crucially, the epistemic value term is positive in the FEF and negative in the EFE, meaning that the FEF penalizes epistemic behavior whereas the EFE promotes it:

\begin{align*}
    \label{FEFEFEComparison}
    \mathbf{FEF} &= -\underbrace{\mathbb{E}_{Q(o_\tau,x_\tau | \pi)} \big[ \ln \tilde{p}(o_\tau | x_\tau) \big]}_{\text{Extrinsic Value}} + \underbrace{\mathbb{E}_{Q(o_\tau | \pi)}\mathbf{D}_{KL}[Q(x_\tau | o_\tau)||Q(x_\tau | \pi)]}_{\text{Epistemic Value}} \numberthis \\
    \mathbf{EFE} &= \underbrace{-\mathbb{E}_{Q(o_\tau,x_\tau | \pi)}\big[ \ln \tilde{p}(o_\tau) \big]}_{\text{Extrinsic Value}} -  \underbrace{\mathbb{E}_{Q(o_\tau | \pi)}\mathbf{D}_{KL}[Q(x_\tau | o_\tau)||Q(x_\tau | \pi)]}_{\text{Epistemic Value}}
\end{align*}

Equation \ref{FEFEFEComparison}. demonstrates that the FEF and EFE can be decomposed in similar fashion. We note that the extrinsic value term for the FEF is a likelihood and a marginal for the EFE. The most important difference, however, lies in the sign of the epistemic value term. Since optimizing either the FEF or the EFE requires their minimization, minimizing the FEF mandates us to minimize information gain while the EFE requires us to maximize it. An FEF agent thus tries to maximize its extrinsic value while trying to explore as little as possible. A key question then arises: where does the negative information gain in the EFE come from?

While this difference in the sign of the expected information gain term may speak to some deep connection between the two quantities, here we offer a pragmatic perspective on the matter. We show that a possible route to the EFE is simply that it is the FEF minus the expected information-gain. This
implies that the epistemic value term of the EFE arises not from some connection to variational inference but is present by construction:
\begin{align*}
    \mathbf{FEF}_\tau(\pi) - \mathbf{IG}_\tau &= \mathbb{E}_{Q(o_\tau,x_\tau | \pi)}\ln(\frac{Q(x_\tau | o_\tau)}{\tilde{p}(o_\tau,x_\tau)}) - \mathbb{E}_{Q(o_\tau,x_\tau | \pi)}\ln(\frac{Q(x_\tau | o_\tau)}{Q(x_\tau | \pi)}) \\
    &=  \mathbb{E}_{Q(o_\tau,x_\tau | \pi)}\ln(\frac{Q(x_\tau|o_\tau)Q(x_\tau | \pi)}{\tilde{p}(o_\tau,x_\tau)Q(x_\tau|o_\tau)}) \\ 
    &=  \mathbb{E}_{Q(o_\tau,x_\tau | \pi)}\ln(\frac{Q(x_\tau | \pi)}{\tilde{p}(o_\tau,x_\tau)}) \\
    &= \mathbf{EFE}(\pi)_\tau
\end{align*}

While this proof illustrates the relation between the EFE and the FEF, it is theoretically unsatisfying as an account of the origin of the EFE. A large part of the appeal of the EFE is that it purports to show that epistemic value arises `naturally' out of minimizing free-energy into the future. In contrast, here we have shown that minimizing free-energy into the future requires no commitment to exploratory behaviour. While this does not question the usefulness of using an information gain term for exploration, or the use of the EFE as a loss function, it does raise questions about the mathematically principled nature of the objective. It is thus not straightforward to see why agents are directly mandated by the FEP to minimize the EFE specifically, as opposed to some other free-energy functional.
While this fact may at first appear concerning, we believe it ultimately enhances the power of the formalism by licensing the extension of active inference to encompass other objective functions in a principled manner \citep{biehl2018expanding}. In the following section, we propose an alternative objective to the EFE, which results in the same information-seeking epistemic value term, but derives it in a mathematically principled and intuitive way as a bound on the divergence between expected and desired futures.

\section{Free Energy of the Expected Future}

In this section, we propose a novel objective functional which we call \textit{The Free-Energy of The Expected Future} (FEEF) which possesses the same epistemic value term as the EFE, while additionally possessing a more naturalistic and intuitive grounding. We begin with the intuition that, to act adaptively, agents should act so as to minimize the difference between what they predict will happen, and what they desire to happen. Put another way, adaptive action for an agent consists of forcing reality to unfold according to its' preferences. We can mathematically formulate this objective as the KL divergence between the agent's veridicial generative model of what is likely to happen, and a biased generative model of what it desires to happen. 
\begin{align*}
    \pi^* = \underset{\pi}{\mathrm{argmin}} \, \, \mathbf{D}_{KL}[Q(o_{t:T},x_{t:T}|\pi)||\tilde{p}(o_{t:T},x_{t:T})]
\end{align*}

The FEEF can be interpreted as the divergence between a veridicial and a biased generative model, and thus furnishes a direct intuition of the goals of a FEEF-minimizing agent. The divergence objective compels the agent to bring the biased and the veridicial generative model into alignment. Since the predictions of the biased generative model are heavily biased towards the agent's a-priori preferences, the only way to achieve this alignment is to act so as to make the veridicial generative model predict desired outcomes in line with the biased generative model. The FEEF objective encompasses the standard active inference intuition of an agent acting through biased inference to maximize accuracy of a biased model. However, the maintenance of two separate generative models (one biased and one veridicial) also helps finesse the conceptual difficulty of how the agent manages to make accurate posterior inferences and future predictions about complex dynamics if all it has access to is a biased generative model. It seems straightforward that the biased model would also bias these crucial parts of inference which need to be unimpaired for the scheme to function at all. However, by keeping both a veridicial generative model (the same one used at the present time and learnt through environmental interactions), and a biased generative model (created by systematically biasing a temporary copy of the veridicial model), we elegantly separate the need for both veridicial and biased inferential components for future prediction \footnote{This approach bears a resemblance to that taken in \citep{friston2019physics} which separates the evolving dynamical policy-dependent density of the agent, and a desired steady state density which is policy-invariant. This approach arises from deep thermodynamic considerations in continuous time, while ours is applicable to discrete time reinforcement learning frameworks.}.

Similarly to the EFE, the FEEF objective can be decomposed into an extrinsic and an intrinsic term. We compare this directly to the EFE decomposition:
\begin{align*}
    \mathbf{FEEF}(\pi)_\tau &= \mathbb{E}_{Q(o_\tau,x_\tau|\pi)} \ln \big[ \frac{ Q(o_\tau,x_\tau |\pi)}{\tilde{p}(o_\tau,x_\tau )} \big] \\
    &= \underbrace{\mathbb{E}_{Q(x_\tau | \pi)} \mathbf{D}_{KL} \big[ Q(o_\tau | x_\tau) \Vert \tilde{p}(o_\tau) \big]}_{\text{Extrinsic Value}} - \underbrace{\mathbb{E}_{Q(o_\tau | \pi)} \mathbf{D}_{KL} \big[ Q(x_\tau | o_\tau) \Vert Q(x_\tau | \pi) \big]}_{\text{Intrinsic Value}} \\
    \mathbf{EFE} &= \underbrace{-\mathbb{E}_{Q(o_\tau,x_\tau | \pi)}\big[ \ln \tilde{p}(o_\tau) \big]}_{\text{Extrinsic Value}} -  \underbrace{\mathbb{E}_{Q(o_\tau | \pi)}\mathbf{D}_{KL}[Q(x_\tau | o_\tau)||Q(x_\tau | \pi)]}_{\text{Intrinsic Value}}
\end{align*}

The first thing to note is that the intrinsic value terms of the FEEF and the EFE are identical, under the assumption that the variational posterior is approximately correct $Q(x_\tau | o_\tau) \approx p(x_\tau | o_\tau)$ such that FEEF-minimizing agents will necessarily show identical epistemic behaviour to EFE-minimizing agents. Unlike the EFE, however, the FEEF also possesses  a strong naturalistic grounding as a bound on a theoretically relevant quantity. The FEEF can maintain both its information-maximizing imperative and its theoretical grounding since it is derived from the minimization of a KL divergence rather than the maximization of a log model evidence. 

The key difference with the EFE lies in the likelihood term. While the EFE simply tries to maximize the expected evidence of the desired observations, the FEEF minimizes the KL divergence between the likelihood of observations predicted under the veridicial generative model \footnote{The term `veridicial' needs some contextualising. We simply mean that the model is not biased towards the agent's desires. The veridicial generative model is not required to be a perfectly accurate map of the agent's entire world, only of action-relevant sub-manifolds of the total space \citep{tschantz2019learning}. } and the marginal likelihood of observations under the biased generative model. This difference is effectively equivalent to an additional veridicial generative model expected likelihood entropy term $\mathbf{H}[Q(o_\tau | x_\tau)]$ subtracted from the EFE. The extrinsic value term thus encourages the agent to choose its actions such that its predictions over states lead to observations which are close to its preferred observations, while also trying move to states whereby the entropy over observations is maximized, thus leading the agent to move towards states where the generative model is not as certain about the likely outcome. In effect, the FEEF possesses another exploratory term, in addition to the information gain, which the EFE lacks. 

Another important advantage of the FEEF is that it is mathematically equivalent to the VFE (with a biased generative model) in the present time with a current observation. This is because when we have a real observation, the distribution over the possible veridicial observations collapses to a delta distribution, so that the outer expectation has no effect as $\mathbb{E}_{Q(o_\tau, x_\tau | \pi)} = \int Q(x_\tau | o_\tau)Q(o_\tau | \pi) = \int Q(x_\tau | o_\tau)\delta(o - \bar{o}) = \int Q(x_\tau | \bar{o}_\tau)$ when a real observation $\bar{o}$ is available. Similarly, the veridicial model can be factorised as $Q(o_\tau, x_\tau) = Q(x_\tau | o_\tau)Q(o_\tau)$ and when the observation is known the entropy of the observation marginal $Q(o_\tau | \pi)$ is 0, thus resulting in the VFE. Simultaneously, biased likelihood is equivalent to the veridicial likelihood $\tilde{p}(\bar{o}_\tau | x_\tau) = Q(\bar{o}_\tau | x_\tau)$, assuming that (barring counterfactual reasoning capability), one cannot usefully desire things to be other than how they are at the present moment. This means that theoretically we can consider an agent to be both inferring and planning using the same objective, which is not true of the EFE. The EFE does not reduce to the VFE when observations are known, and thus requires a separate objective function to be minimized for planning compared to inference. Because of this, it is actually possible to argue that FEEF is mandated by the free-energy principle. On this view there is no distinction between present and future inference and both follow from  minimizing the same objective but under  different informational constraints.

Since the FEEF and the EFE are identical in their intrinsic value term, and share deep similarities in their extrinsic term, we believe that the FEEF can serve as a relatively straightforward "plug-in replacement" for the EFE for many active inference agents. Moreover, it has a much more straightforward intuitive basis than the EFE, is arguably a better continuation of the VFE into the future, and possesses a strong naturalistic grounding as a bound on the divergence between predicted and desired futures. 

\section{Discussion}

We believe it is valuable at this point to step back from the morass of various free-energies and take stock of what has been achieved. Firstly, we have shown that it is not possible to directly derive epistemic value from variational inference objectives which serve as a bound on model evidence. However, it is possible to derive epistemic value terms from divergences between the biased and veridicial generative models. A deep intuitive understanding of why this is the case is an interesting avenue for future work. The intuition behind the FEEF as a divergence between desired and expected future observations is also similar to probabilistic formulations of the reinforcement learning problem \citep{levine2018reinforcement,attias2003planning,toussaint2009probabilistic,kappen2005path}, which typically try to minimize the divergence between a controlled trajectory, and an optimal trajectory \citep{kappen2007introduction,theodorou2012relative,williams2017model}. These schemes also obtain some degree of (undirected) exploratory behaviour through their objective functionals which contain entropy terms and the FEEF can be seen as a way of extending these schemes to partially-observed environments. Understanding precisely how active inference and the free-energy principle relate mathematically to such schemes is another fruitful avenue for future work.

It seems intuitive that a Bayes-optimal solution to the exploration-exploitation dilemma should arise directly out of the formulation of reward maximization as inference, given that sources of uncertainty are correctly quantified. However, in this paper, we have shown that merely quantifying uncertainty in states and observations through mean-field-factorised time-steps is insufficient to derive such a principled solution to the dilemma, as seen by the exploration-discouraging behaviour of the FEF. We therefore believe that to derive Bayes optimal exploration policies in the context of active-learning -- such that we have to select actions that give us the most information now to use in the future to maximize rewards -- it is likely to require both modelling multiple interconnected time-steps, as well as the mechanics of learning with parameters and update rules, and correctly quantifying the uncertainties therein. This is beyond the scope of this paper, but is a very interesting avenue for future work. 

The comparison of the FEEF and the EFE also raises an interesting philosophical point about the number and types of generative models employed in the active-inference formalism. One interpretation of the FEEF is in terms of two generative models, but other interpretations are possible such as between a single unbiased generative model and a simple density of desired states and observations. It is also important to note that due to requiring different objective functions for inference and planning, the EFE also formulation appears to implicitly require two generative models -- the generative model of future states, and the generative model of states in the future \citep{friston2015active}. While the mathematical formalism is relatively straightforward, the philosophical question of how to translate the mathematical objects into ontological objects called `generative models' is unclear and progress on this front would be useful in determining the philosophical status, and perhaps even neural implementation of active inference.

The implications of our results for studies of active inference are varied. Nothing in what we have shown argues directly against the use of the EFE as a objective for an active inference agent. However, we believe we have shown that the EFE is not the necessarily the only, or even the natural, objective function to use. We thus follow \citep{biehl2018expanding} in encouraging experimentation with different objective functions for active inference. We especially believe that our objective, the FEEF future has promise due its intuitive interpretation, largely equivalent terms to the EFE, its straightforward use of two generative models rather than just a single biased one, and its close connections to similar probabilistic objectives used in variational reinforcement learning, while also maintaining the crucial epistemic properties of the EFE.  Moreover, while in this paper we have argued for the FEF instead of the EFE as a direct extension of the VFE into the future, the logical requirements of exactly which functional (if any) is, in fact, mandated by the free-energy principle remains open. We believe that elucidating the exact constraints which the free-energy principle places upon a theory of variational action, and understanding more deeply the relations between the various free-energies, could shed light on deep questions regarding notions of Bayes-optimal epistemic action in self-organising systems.

Finally, it is important to note that although in this paper we have solely been concerned with the EFE and active inference in discrete-time POMDPs, the original intuitions and mathematical framework of the free-energy principle arose out of a continuous time formulation, deeply interwoven with concerns from information theory and statistical physics \citep{friston2006free,friston2012ao,friston2019physics,parr2020markov}. As such there may be deep connections between the EFE, FEF, and log model evidence which exist only in the continuous time limit, and which furnish a mathematically principled origin of epistemic action. 

\section{Conclusion}

In this paper, we have examined in detail the nature and origin of the EFE. We have shown that it is not a direct analog of the VFE extended into the future. We then derived a novel objective, the FEF, which we claimed is a more natural extension and shown that it lacks the beneficial epistemic value term of the EFE. We then proved that this term arises in the EFE directly as a result of its non-standard definition since the EFE can be expressed as just the FEF minus the expected information gain. Taking this into account, we then proposed another objective, the \textit{Free Energy of the Expected Future} (FEEF) which attempts to get the best of both worlds by preserving the desirable information-seeking properties of the EFE, while also maintaining a mathematically principled origin.

\section{Acknowledgements}
BM is supported by an EPSRC
funded PhD Studentship. AT is funded by a PhD studentship from the Dr. Mortimer and Theresa Sackler Foundation and the
School of Engineering and Informatics at the University of Sussex. CLB is supported by BBRSC grant number BB/P022197/1. AT is grateful to the Dr. Mortimer and Theresa Sackler Foundation, which supports the Sackler Centre for Consciousness Science. 

\bibliography{NECO-20-001-35R1-SUPPPLEMENTARY.bib}

\section{Variational Inference}
\label{variational}

To motivate the variational free-energy, and variational inference more generally, we setup a standard inference problem. Let us say we are an agent that exists in a partially observed world. We have some observation $o_t$, and from this we wish to infer the hidden state of the world $x_t$. That is, we want to compute the posterior $p(x_t|o_t)$. While we do not know this posterior directly, we do possess a generative model of the world. This is a model that maps from hidden states to observations. Mathematically, we possess $p(o_t,x_t) = p(o_t|x_t)p(x_t)$. Since computing the true posterior exactly is likely intractable, the strategy in variational inference is to try to approximate this density with a tractable one $Q(x_t | o_t; \phi)$ which we postulate, and thus have full control over. While the true posterior might be arbitrarily complex, we might define $Q(x_t | o_t; \phi)$ to be a Gaussian distribution: $Q(x_t | o_t; \phi) = \mathcal{N}(x; \mu_\phi, \sigma_\phi)$, for instance. Given that we have this variational density \textit{q}, parametrised by some parameters $\phi$, the goal is to adjust the parameters to make \textit{q} as close as possible to the true posterior $p(x_t|o_t)$. Mathematically speaking, this means we want to minimize:
\begin{align*}
    \mathrm{arg min}_{\phi} \mathbf{D}_{KL}[Q(x_t | o_t; \phi)||p(x_t|o_t)]
\end{align*}
Where $\mathbf{D}_{KL}[Q \Vert P]$ is the Kullback-Leibler divergence. This initially doesn't seem to have bought us much. We wish to minimize the divergence between the variational density q and the true posterior $p(x_t|o_t)$. However, by assumption, we do not know the true posterior. So how can we possibly minimize this divergence if we do not know one of the parts? This is where we use the key trick of variational inference. By Bayes theorem we know that: $p(x_t|o_t) = \frac{p(o_t | x_t)p(x_t)}{p(o_t)}$ we we can thus substitute this into the KL divergence term.
\begin{align*}
    \mathrm{arg min}_{\phi} \mathbf{D}_{KL}[Q(x_t | o_t; \phi)||p(x_t|o_t)] &= \mathrm{arg min}_{\phi} \mathbf{D}_{KL}[Q(x_t | o_t; \phi)||\frac{p(o_t | x_t)p(x_t)}{p(o_t)}] \\
    &= \mathbb{E}_{Q(x_t | o_t; \phi)}\ln (\frac{Q(x_t | o_t; \phi)p(o_t)}{p(o_t | x_t)p(x_t)}) \\
    &= \mathbb{E}_{Q(x_t | o_t; \phi)}\ln (\frac{Q(x_t | o_t; \phi)}{p(o_t | x_t)p(x_t)}) + \mathbb{E}_{Q(x_t | o_t; \phi)}\ln p(o_t) \\
    &= \mathbf{D}_{KL}[Q(x_t | o_t; \phi)||p(o_t | x_t)p(x_t)] + \ln p(o_t) \numberthis \label{VFE_deriv}
\end{align*}

In step 2 we have applied Bayes theorem the the posterior. In step 3 we have simply utilized the definition of the KL-divergence $\mathbf{D}_{KL}[Q||P] = \mathbb{E}_Q \ln(\frac{Q}{P})$. In step 4 we have then applied the property of logs that $\ln (a * b) = \ln(a) + \ln(b)$. In step 5 we then recognise that the remaining first term is now a KL divergence between the variational posterior and the generative model. We also recognise that since the $\ln p(o_t)$ term has no dependence on $x$ or $\phi$, the expectation $ \mathbb{E}_{Q(x_t | o_t; \phi)}\ln p(o_t)$ vanishes leaving just the $\ln p(o_t)$ term alone. It is important to note that the KL term in equation \ref{VFE_deriv} is now between two things we can actually compute -- the variational posterior, which we control, and the generative model, which we assume that we know. The remaining $\ln p(o_t)$ term is called the log model evidence and it is incomputable in general. However, since it is not affected by the parameters $\phi$ of the variational density, then it does not affect the minimization and so for the purposes of the minimization process can be ignored. We can thus write out what we have defined as 
\begin{align*}
     & \mathbf{D}_{KL}[Q(x_t | o_t; \phi)||p(x_t|o_t)] = \mathbf{D}_{KL}[Q(x_t | o_t; \phi)||p(o_t | x_t)p(x_t)] + \ln p(o_t) \\
     &\Rightarrow \mathbf{D}_{KL}[Q(x_t | o_t; \phi)||p(o_t | x_t)p(x_t)] \geq  \mathbf{D}_{KL}[Q(x_t | o_t; \phi)||p(x_t|o_t)] 
\end{align*}
This implies that the KL divergence between the variational density and the generative model is always greater than or equal to the KL divergence between the true and variational posteriors. Since we can compute the first KL divergence, we call it the variational free-energy $F$. Since it is an upper bound on the divergence between the true posterior and the variational posterior, which is what we really want to minimize, then if we minimize $F$, we are constantly pushing that bound lower and thus largely minimizing the divergence between the true and variational posterior. As an additional bonus, when the true and variational posteriors are approximately equal: $ \mathbf{D}_{KL}[Q(x_t | o_t; \phi)||p(x_t|o_t)]  \approx 0$ then $\mathbf{D}_{KL}[Q(x_t | o_t; \phi)||p(o_t | x_t)p(x_t)] \approx -\ln p(o_t)$, which means that the final value of the variational-free-energy is thus equal to the negative log model evidence. Since the log model evidence is a very useful quantity to compute for Bayesian model selection, it effectively means that once we have finished fitting our model, we are automatically left with a measure of how good our model is.

In effect the variational free energy is useful because it has two properties. The first is that it is an upper bound on the divergence between the true and approximate posterior. By adjusting our approximate posterior to minimize this bound, we drive it closer to the true posterior, thus achieving more accurate inference. Secondly, the variational free-energy is a bound on the log model evidence. This is an important term which scores the likelihood of the data observed given your model and so can be used in Bayesian model selection. 

The log model evidence takes on an additional importance in terms of the free-energy principle, since the negative log model evidence $-\ln p(o_t)$ is surprisal, which all agents, it is propsed are driven to minimize \citep{friston2006free}. This is because the expected log model evidence is the entropy of observations, the minimisation of which is postulated as a necessary condition for any self-sustaining organism to maintain itself as a unique system. The free-enregy minimization comes about since the VFE is, as we have seen a tractable bound on the log model evidence, or surprisal.

The VFE can be decomposed in three principle ways, which each showcases a different facet of the objective.

\begin{align*}
    \mathbb{F} & = \mathbf{D}_{KL}[Q(x_t | o_t; \phi)||p(o_t,x_t)] \\
    &= \mathbb{E}_{Q(x_t | o_t; \phi)} \big[ \ln \frac{Q(x_t | o_t; \phi)}{p(o_t,x_t)} \big] \\
    &= \underbrace{\mathbb{E}_{Q(x_t | o_t; \phi)}[\ln Q(x_t | o_t; \phi)]}_{\text{Entropy}} - \underbrace{\mathbb{E}_{Q(x_t | o_t; \phi)}[\ln {p(o_t,x_t)}]}_{\text{Energy}} \\
    &= \underbrace{-\mathbb{E}_{Q(x_t | o_t; \phi)}[\ln p(o_t | x_t)]}_{\text{Accuracy}} + \underbrace{\mathbf{D}_{KL}[Q(x_t | o_t; \phi)||p(x_t)]}_{\text{Complexity}} \\
    &= \underbrace{-\ln p(o_t)}_{\text{Negative Log Model Evidence}} + \underbrace{\mathbf{D}_{KL}[Q(x_t | o_t; \phi)||p(x_t | o_t)]}_{\text{Posterior Divergence}}
\end{align*}

In the first entropy-energy decomposition, we simply split the KL divergence using the properties of logarithms so that the numerator of the fraction becomes the entropy term and the denominator becomes the energy term. If we are seeking to minimize the variational-free-energy then this means we need to both minimize the negative entropy (since entropy is defined as $-\mathbb{E}_{Q(x)}\big[ \ln Q(x) \big]$ and also minimize the negative energy (or maximize the energy) $\mathbb{E}_{Q(x_t | o_t; \phi)}[\ln {p(o_t,x_t)}]$. This can be interpreted as saying we require that the variational posterior be as entropic as possible while also maximizing the likelihood that the $x$s proposed as probable by the variational posterior also be judged as probable under the generative model.

The second decomposition into accuracy and complexity perhaps has a more straightforward interpretation. We wish to minimize the negative accuracy (and thus maximize the accuracy), which means we want the actually observed observation to be as likely as possible under the $xs$ predicted by the variational posterior. However, we also want to minimize the complexity term which is a KL divergence between the variational posterior and the prior. That is, we wish to keep your posterior as close to our prior as possible while still maximizing accuracy. The complexity term then functions as a kind of implicit regulariser, making sure we do not overfit to any specific observation.

The final decomposition speaks the the inferential functions of the VFE. It serves as an upper bound on the log model evidence, since the posterior divergence term, as a KL divergence, is always positive. Moreover, we see that by minimizing the free-energy, we must also be minimizing the posterior divergence, which is the difference between the approximate and true posterior, and we are thus improving our variational approximation.

\section{Decompositions of the EFE}
\label{decompositions}

In this section we provide a comprehensive overview of the many decompositions of the EFE. The EFE is defined as:
\begin{align*}
     \mathcal{G}(\pi) = \mathbb{E}_{Q(o_\tau,x_\tau | \pi)}[\ln Q(x_\tau | \pi) - \ln \tilde{p}(o_\tau,x_\tau)]
\end{align*}

The standard decomposition is into the extrinsic term (expected log likelihood of the desired observations) and an epistemic term (the information gain, or KL divergence between variational prior and posterior from the generative model.
\begin{align*}
   \mathbb{E}_{Q(o_\tau,x_\tau | \pi)}[\ln Q(x_\tau | \pi) - \ln \tilde{p}(o_\tau,x_\tau)] &= \mathbb{E}_{Q(o_\tau,x_\tau | \pi)}[- \ln \tilde{p}(o_\tau) - \ln p(x_\tau |o_\tau) + \ln Q(x_\tau | \pi)] \\
   &= \underbrace{-\mathbb{E}_{Q(o_\tau,x_\tau | \pi)}\big[ \ln \tilde{p}(o_\tau) \big]}_{\text{Extrinsic Value}} -  \underbrace{\mathbb{E}_{Q(o_\tau | \pi)} \big[ \mathbf{D}_{KL}[Q(x_\tau | o_\tau)||Q(x_\tau | \pi)] \big]}_{\text{Epistemic Value}} 
\end{align*}

Similar to the VFE, it is also possible to split it into an energy and an entropy term. While the energy term is similar to the VFE as the expectation of the generative model (albeit an expectation over the joint instead of the posterior), the entropy term is different as it is the entropy of the variational \textit{prior}, not the approximate posterior, which results.
\begin{align*}
    \mathcal{G}(\pi) &= \mathbb{E}_{Q(o_\tau,x_\tau | \pi)}[\ln Q(x_\tau | \pi) - \ln \tilde{p}(o_\tau,x_\tau)] \\
    &= \mathbb{E}_{Q(o_\tau,x_\tau | \pi)}[\ln Q(x_\tau | \pi)] - \mathbb{E}_{Q(o_\tau,x_\tau | \pi)}[\ln \tilde{p}(o_\tau,x_\tau)] \\
    &= -\underbrace{\mathbb{E}_{Q(o_\tau| x_\tau)} \big[ \mathcal{H}\big[Q(x_\tau | \pi) \big] \big]}_{\text{Entropy}} -  \underbrace{\mathbb{E}_{Q(o_\tau,x_\tau | \pi)}[\ln \tilde{p}(o_\tau,x_\tau)]}_{\text{Energy}}
\end{align*}
It is also possible to decompose the biased generative model the other way around, thus in line with that of the VFE to derive:
\begin{align*}
    \mathcal{G}(\pi) &= \mathbb{E}_{Q(o_\tau,x_\tau| \pi)}[\ln Q(x_\tau | \pi) - \ln \tilde{p}(o_\tau,x_\tau)] \\
    &= \mathbb{E}_{Q(o_\tau,x_\tau| \pi)}[\ln Q(x_\tau | \pi) - \ln \tilde{p}(o_\tau| x_\tau) - \ln p(x_\tau)] \\
    &=  \underbrace{-\mathbb{E}_{Q(o_\tau,x_\tau | \pi)} \big[ \ln \tilde{p}(o_\tau| x_\tau) \big]}_{\text{Accuracy}} + \underbrace{\mathbb{E}_{Q(o_\tau| x_\tau)} \big[ \mathbf{D}_{KL} \big[ Q(x_\tau | \pi) \Vert p(x_\tau) \big] \big]}_{\text{Complexity}}
\end{align*}

Unlike the VFE, however the divergence is between the variational prior and the generative prior, rather than between the variational posterior and the generative prior. Finally, the EFE can also be represented in observation space by using Bayes rule to flip the likelihoods and priors. 
\begin{align*} %
    \mathcal{G}(\pi) &= \mathbb{E}_{Q(o_\tau,x_\tau | \pi)}[\ln Q(x_\tau | \pi) - \ln \tilde{p}(o_\tau,x_\tau)] \\
    &= \mathbb{E}_{Q(o_\tau,x_\tau | \pi)}[\ln Q(x_\tau | \pi) - \ln \tilde{p}(o_\tau) - \ln Q(x_\tau | o_\tau)] \\
    &= \mathbb{E}_{Q(o_\tau,x_\tau | \pi)}[\ln Q(x_\tau | \pi) - \ln \tilde{p}(o_\tau) - \ln Q(o_\tau | x_\tau) - \ln Q(x_\tau | \pi) + \ln Q(o_\tau)] \\
    &= \mathbb{E}_{Q(o_\tau,x_\tau | \pi)}[- \ln \tilde{p}(o_\tau) - \ln Q(o_\tau | x_\tau) + \ln Q(o_\tau)] \\
    &= \underbrace{\mathbb{E}_{Q(x_\tau | \pi)}\big[ \mathcal{H}\big[ Q(o_\tau| x_\tau) \big] \big]}_{\text{Predicted Uncertainty}} - \underbrace{\mathbb{E}_{Q(x_\tau | o_\tau)} \big[ \mathbf{D}_{KL} \big[ Q(o_\tau) \Vert \tilde{p}(o_\tau) \big] \big]}_{\text{Predicted Divergence}} \\
    &= - \underbrace{\mathbb{E}_{Q(x_\tau | \pi)}\big[ \ln \tilde{p}(o_\tau) \big]}_{\text{Extrinsic Value}} - \underbrace{\mathbb{E}_{Q(x_\tau | \pi)} \big[ \mathbf{D}_{KL} \big[ Q(o_\tau | x_\tau) \Vert Q(o_\tau) \big] \big]}_{\text{(Observation) Information Gain}}
\end{align*} %
It is also possible to factorise the biased generative model the other way around in terms of an unbiased likelihood and biased states: $\tilde{p}(o_\tau, x_\tau) = p(o_\tau | x_\tau)\tilde{p}(x_\tau)$. This different factorisation leads to a new decomposition in terms of risk and ambiguity, as well as potentially different behaviour due to the change from desired observations to desired states \footnote{For further detail on this factorization see \citet{da2020active}.}.
\begin{align*}
    \mathcal{G}(\pi) &= \mathbb{E}_{Q(o_\tau,x_\tau | \pi)}[\ln Q(x_\tau | \pi) - \ln \tilde{p}(o_\tau,x_\tau)] \\
    &= \mathbb{E}_{Q(o_\tau,x_\tau | \pi)}[\ln Q(x_\tau | \pi) - \ln p(o_\tau | x_\tau) - \ln \tilde{p}(x_\tau)] \\
    &= \underbrace{\mathbb{E}_{Q(x_\tau | \pi)} \big[ \mathcal{H}[p(o_\tau | x_\tau)] \big]}_{\text{Ambiguity}} + \underbrace{\mathbf{D}_{KL} \big[ Q(x_\tau | \pi) \Vert\tilde{p}(x_\tau | \pi) \big]}_{\text{Risk}}
\end{align*}
Here the agent is driven to minimize the divergence between desired and prior expected states, while also trying to minimize the entropy of the observations it receives. This drives the agent to try to sample observations with a minimally ambiguous (or maximally precise) mapping back to states.

This formulation is mathematically equivalent to the previous decompositions despite defining desired states instead of desired observations, as can be seen with the following manipulations:
\begin{align*}
    \mathcal{G}(\pi) &=  \underbrace{\mathbb{E}_{Q(x_\tau | \pi)}\big[ \mathcal{H}[p(o_\tau | x_\tau)] \big]}_{\text{Ambiguity}} + \underbrace{\mathbf{D}_{KL} \big[ Q(x_\tau) \Vert\tilde{p}(x_\tau) \big]}_{\text{Risk}} \\
    &= \mathbb{E}_{Q(o_\tau,x_\tau | \pi)}[\ln Q(x_\tau | \pi) - \ln p(o_\tau | x_\tau) -\ln Q(x_\tau | o_\tau) - \ln \tilde{p}(o_\tau) + \ln p(o_\tau | x_\tau)] \\
    &= \underbrace{-\mathbb{E}_{Q(o_\tau,x_\tau | \pi)}\big[ \ln \tilde{p}(o_\tau) \big]}_{\text{Extrinsic Value}} -  \underbrace{\mathbb{E}_{Q(o_\tau | \pi)}\big[ \mathbf{D}_{KL}[Q(x_\tau | o_\tau)||Q(x_\tau | \pi)] \big]}_{\text{Epistemic Value}} = \mathcal{G}(\pi)
\end{align*}
The risk-ambiguity formulation has very close relations to KL control \citep{rawlik2013stochastic}, in that it encompasses KL control with an additional "epistemic" ambiguity term.
\begin{align*}
    \mathcal{G}(\pi) &=  \underbrace{\mathbb{E}_{Q(x_\tau | \pi)}\mathcal{H}[p(o_\tau | x_\tau)] + \underbrace{\mathbf{D}_{KL} \big[ Q(x_\tau | \pi) \Vert\tilde{p}(x_\tau) \big]}_{\text{KL Control}}}_{\text{Active Inference}} \\
\end{align*}
\section{Trajectory Derivation of the Expected Model Evidence}
\label{trajectory}

Here we present the derivation of the free energy of the future (FEF) from the expected model evidence for the full trajectory distribution rather than a single time-step. Importantly, we show that with a temporal mean-field approximation on the approximate posterior: $p(x_{1:T}|o_{1:T}) \approx \prod_t^T p(x_t | o_t)$, the assumption that desired rewards are independent in time: $p(\hat{r}_{1:T}) \approx \prod_t^T p(\hat{r}_t)$, and given a Markovian generative model, then the trajectory distribution factorises into a sum of individual time-steps \footnote{We assume discrete time so there is a sum over timesteps. We also assume continuous states so there is an integral over states $x$. However, the derivation is identical in the case of discrete states where the integral is simply replaced with a sum. }, only dependent on the past through the prior term $p(x_t) = \mathbb{E}_{Q(x_{t-1}|o_{t-1}}p(x_t | x_{t-1})$. We name this final approximation the factorization approximation, and it simply states that your prior at the current time-step is based on the posterior of the previous time-step mapped through the transition dynamics $p(x_t | x_{t-1})$.

\begin{align*}
    & \mathrm{arg min}_{p(\pi)} - \mathbb{E}_{Q(o_{1:T} | \pi)}\ln \tilde{p}(o_{1:T}) \\
    &= -\mathbb{E}_{Q(o_{1:T} | \pi)}\ln \int dx_{1:T} \, \tilde{p}(o_{1:T},x_{1:T}) \\
    &= -\mathbb{E}_{Q(o_{1:T} | \pi)}\ln \int dx_{1:T}  \, \frac{\tilde{p}(o_{1:T},x_{1:T})Q(x_{1:T}|o_{1:T})}{Q(x_{1:T}|o_{1:T})} \\
    &= -\mathbb{E}_{Q(o_{1:T} | \pi)}\ln \int dx_{1:T} \, \prod_t^T \frac{\tilde{p}(o_t,x_t)Q(x_t|o_t)}{Q(x_t|o_t)} \\
    &= -\mathbb{E}_{Q(o_{1:T} | \pi)}\ln \int dx_{1:T}\,  \prod_t^T \frac{ \tilde{p}(o_t | x_t)\mathbb{E}_{Q(x_{t-1}|o_{t-1})}p(x_t | x_{t-1}) Q(x_t | o_t)}{Q(x_t | o_t)} \\
    &= \mathbb{E}_{Q(o_{1:T} | \pi)}\sum_t^t \ln \int dx_t \,  \frac{ \tilde{p}(o_t | x_t)-\mathbb{E}_{Q(x_{t-1}|o_{t-1})}p(x_t | x_{t-1}) Q(x_t | o_t)}{Q(x_t | o_t)} \\
    &\geq -\sum_t^t  \mathbb{E}_{Q(o_{1:T} | \pi)} \int dx_t \, Q(x_t | o_t) \ln \frac{ \tilde{p}(o_t | x_t)\mathbb{E}_{Q(x_{t-1}|r_{t-1})}p(x_t | x_{t-1})}{Q(x_t | o_t)} \\
    &\geq -\sum_t^t  \int dx_t \,  \int do_{1:T} \, Q(o_{1:T},x_t | \pi) \ln \frac{ \tilde{p}(o_t | x_t)\mathbb{E}_{Q(x_{t-1}|o_{t-1})}p(x_t | x_{t-1})}{Q(x_t | o_t)} \\
    &\geq -\sum_t^t  \mathbb{E}_{Q(o_t,x_t | \pi)} \ln \frac{ \tilde{p}(o_t | x_t)\mathbb{E}_{Q(x_{t-1}|o_{t-1})}p(x_t | x_{t-1})}{Q(x_t | o_t)} \\
    &\geq -\sum_t^t  \mathbb{E}_{Q(o_t, x_t | \pi)}\ln \tilde{p}(o_t | x_t) -  \sum_t^t  \mathbb{E}_{p(o_t)}\mathbf{D}_{KL}[p(x_t | o_t)||\mathbb{E}_{Q(x_{t-1}|o_{t-1})}p(x_t | x_{t-1})] \\
    &\geq -\sum_t^t \mathbf{FEF}_t
\end{align*}

The trajectory derivation of the FEEF follows an almost identical scheme to that of the FEF. The only difference is that now the term inside the log also contains an additional $-\ln \tilde{p}(o)$, which is then combined with the likelihood from the generative model to form the extrinsic-value KL divergence.

\section{EFE Bound on the Negative Log Model Evidence}
\label{bound}

It is important to note that the EFE is also a bound on the negative log model evidence, but a lower bound, not an upper bound. This means that in theory, one should want to \textit{maximize} the EFE, instead of minimize it, to make the bound as tight as possible.

It is straightforward to show the bound, since the extrinsic value term of the EFE simply \textit{is} the log model evidence.
\begin{align*}
    \mathbf{EFE} &= \mathbb{E}_{Q(o_\tau,x_\tau | \pi)} [\ln Q(x_\tau | \pi) - \ln \tilde{p}(o_\tau,x_\tau)] \\
    &\approx \mathbb{E}_{Q(o_\tau,x_\tau | \pi)} [\ln Q(x_\tau | \pi) - \ln Q(x_\tau |o_\tau) - \ln \tilde{p}(o_\tau)] \\
    &\approx \underbrace{-\mathbb{E}_{Q(o_\tau | \pi)} [ \ln \tilde{p}(o_\tau)]}_{\text{Negative Expected Log Model Evidence}} -  \underbrace{\mathbb{E}_{Q(o_\tau | \pi)}\mathbf{D}_{KL}[ Q(x_\tau | o_\tau) \Vert Q(x_\tau | \pi)]|}_{\text{Information Gain}}
\end{align*}

This derivation assumes that the true and approximate posteriors are approximately equal $p(x_\tau | o_\tau) \approx Q(x_\tau | o_\tau)$  such that this is true only after a variational inference procedure is completed.

We wish to minimize both log model evidence, and minimize the EFE. Since the information gain term is a KL divergence, which is always $\geq 0$, and we have a negative information gain term, this means that the EFE is always less than the log model evidence and so is a lower bound. However, this bound becomes tight when the information gain is 0, so to maximally tighten the bound we wish to reduce the information gain, while the EFE demands we maximize it. In effect, this means that the EFE bound is the wrong way around.

We can see this more clearly when we retrace the logic for the FEF. From equation \ref{FEF_bound}, we have that the FEF is an \textit{upper bound} on the negative log model evidence. This means that minimizing the FEF necessarily tightens the bound, while this is not true of the EFE \textit{lower bound}, where minimizing the EFE can actually cause it to diverge from the log model evidence. We can see this even more clearly by doing an analogous decomposition of the FEF.
\begin{align*}
    \mathbf{FEF} &= \mathbb{E}_{Q(o_\tau,x_\tau | \pi)} [\ln Q(x_\tau | o_\tau) - \ln \tilde{p}(o_\tau,x_\tau)] \\
    &= \mathbb{E}_{Q(o_\tau,x_\tau | \pi)} [\ln Q(x_\tau | o_\tau) -\ln p(x_\tau|o_\tau) - \ln \tilde{p}(o_\tau)] \\
    &= \underbrace{-\mathbb{E}_{Q(o_\tau | \pi)} [ \ln \tilde{p}(o_\tau)]}_{\text{Negative Expected Log Model Evidence}} +  \underbrace{\mathbb{E}_{Q(o_\tau | \pi)}\mathbf{D}_{KL}[ Q(x_\tau | o_\tau) \Vert p(x_\tau|o_\tau)]|}_{\text{Posterior Approximation Error}}
\end{align*}

Here, since the KL is between the generative model and the approximate posterior, and then decompose the generative model into a true posterior and marginal, we can no longer make the assumption, made in the EFE derivation, that the true and approximate posterior are approximately equal, since that would leave us with only the model evidence. Therefore, instead we get a posterior approximation error term which is the KL divergence between the approximate and true posteriors. When the true and approximate posterior are equal, we are just left with the log model evidence. Since, the posterior approximation error is always $\geq 0$, then the FEF is an upper bound on the negative log model evidence, and thus by minimizing the FEF, we make the bound tighter. This logic is essentially a reprise of the standard variational inference logic from a slightly different perspective. 

If we do not make the assumption in the EFE that the approximate and true posterior are the same, we can derive a similar expression to the EFE which will shed more light on the relation.
\begin{align*}
    \mathbf{EFE} &= \mathbb{E}_{Q(o_\tau,x_\tau | \pi)} [\ln Q(x_\tau | \pi) - \ln \tilde{p}(o_\tau,x_\tau)] \\
    &\approx \mathbb{E}_{Q(o_\tau,x_\tau | \pi)} [\ln Q(x_\tau | \pi) - \ln p(x_\tau |o_\tau) - \ln \tilde{p}(o_\tau)] \\
    &\approx \mathbb{E}_{Q(o_\tau,x_\tau | \pi)} [\ln Q(x_\tau | \pi) - \ln p(x_\tau |o_\tau) - \ln \tilde{p}(o_\tau) +\ln Q(x_\tau |o_\tau) - \ln Q(x_\tau | o_\tau)] \\
    &\approx \underbrace{\underbrace{-\mathbb{E}_{Q(o_\tau | \pi)} [ \ln \tilde{p}(o_\tau)]}_{\text{Negative Expected Log Model Evidence}} +  \underbrace{\mathbb{E}_{Q(o_\tau | \pi)}\mathbf{D}_{KL}[ Q(x_\tau | o_\tau) \Vert p(x_\tau|o_\tau)]|}_{\text{Posterior Approximation Error}}}_{\text{FEF}}  -  \underbrace{\mathbb{E}_{Q(o_\tau | \pi)}\mathbf{D}_{KL}[ Q(x_\tau | o_\tau) \Vert Q(x_\tau | \pi)]|}_{\text{Information Gain}}
\end{align*}

Without the true posterior assumption, we thus find that the EFE could be both an upper or a lower bound on the log model evidence, since the two additional KL divergence terms have opposite signs. If the posterior approximation error is larger than the information gain, then the EFE functions correctly as an upper bound. However, if the information gain is larger, then the EFE will become a lower bound and could diverge from the log model evidence. Moreover, this latter situation is more likely, since the goal of variational inference is to reduce the approximation error, while EFE agents seek to maximize information gain. This means that the EFE only functions correctly as an upper bound on log model evidence during the early stages of optimization where the posterior approximation is poor. Further optimization steps likely drive the EFE further away from the model evidence. The bound is tight when the information gain equals the posterior approximation error. We can also see that the first two terms of the EFE is simply the FEF, we have thus rederived by a rather roundabout route, the fact that the EFE is simply the FEF minus the information gain.

We thus see that the EFE as a bound on the log model evidence is shaky, since it depends on the information gain always being larger or smaller than the posterior approximation error. Moreover, the bounding behaviour seems to emerge directly from the relation of the EFE to the FEF rather than the intrinsic qualities of the EFE, and it is primarily the information-seeking properties of the EFE which serve to damage the clean bounding behaviour of the FEF.

It can be argued that although the mathematical justification of the EFE as a bound may be shaky, that the additional information gain term may be beneficial, and the bound may be recovered in the long run, since that as a result of short-term actions to maximize the EFE, the epistemic value itself goes to 0, and thus the EFE exactly approximates the bound, while also potentially increasing the ultimate expected reward achieved. This argument is valid heuristically and is identical to the standard justifications for ad-hoc intrinsic measures terms in the literature \citep{oudeyer2009intrinsic} namely that exploration hurts in the short run but helps in the long run. We do not dispute that argument in this paper, instead we simply show that the EFE cannot straightforwardly be justified mathematically as  being a result of variational inference into the future, or as a bound on model-evidence. We do not argue at all against its heuristic use to encourage exploration of the environment and thus (hopefully) better performance overall.
\section{Attempts at Naturalising the EFE}
\label{naturalising}

In this appendix, we review several attempts to derive the EFE directly from the expected model evidence. 

Since we have derived the FEF by importance sampling the expected model evidence with the approximate posterior, one obvious avenue would be to importance sample on the variational prior instead. Following this line of thought gives us:
\begin{align*}
    - \mathbb{E}_{Q(o_\tau | \pi)} \big[ \ln \tilde{p}(o_\tau) \big] &= - \mathbb{E}_{Q(o_\tau | \pi)} \big[ \ln \int dx \tilde{p}(o_\tau,x_\tau) \big] \\
    &= - \mathbb{E}_{Q(o_\tau | \pi)} [\big[ \ln \int dx_\tau \tilde{p}(o_\tau,x_\tau)\frac{Q(x_\tau| \pi)}{Q(x_\tau | \pi)} \big] \\
    &\leq - \mathbb{E}_{Q(o_\tau | \pi)} \big[ \int dx_\tau Q(x_\tau | \pi) \ln \frac{\tilde{p}(o_\tau,x_\tau)}{Q(x_\tau | \pi)} \big] \\
    &\leq - \mathbb{E}_{Q(o_\tau | \pi)Q(x_\tau | \pi)} \big[ \ln \frac{\tilde{p}(o_\tau,x_\tau)}{Q(x_\tau | \pi)} \big] \\
    &\leq - \mathbb{E}_{Q(o_\tau | \pi)Q(x_\tau | \pi)} [ \ln Q(x_\tau | \pi) - \ln \tilde{p}(o_\tau,x_\tau)]
\end{align*}

While this approach gets the correct form of the EFE inside the expectation, the expectation itself is the product of the two marginals rather than the joint required for the full EFE. While this may seem minor, this difference must underpin all the other differences and relations we have explored throughout this paper.

To get to the full EFE we must make some assumption to allow us to combine the expectation under two marginals into an expectation under the joint. The first and simplest assumption is that they simply are the same such that the joint factorises into the two marginals -- $Q(o_\tau,x_\tau | \pi) \approx Q(o_\tau | \pi)Q(x_\tau|\pi)$. This assumption is equivalent to assuming independence of observations and latent states, which rather defeats the point of a latent variable model. 

A second approach is to assume that the variational prior equals the variational posterior $Q(x_\tau | \pi) \approx Q(x_\tau | o_\tau)$. This allows you then to combine the marginal and posterior into a joint, giving the EFE as desired. However this assumption has several unfortunate consequences. Firstly, it eliminates the entire idea of inference, since the prior and posterior are assumed to be the same, thus no real inference can have taken place. This is not necessarily an issue if we separate the inference and planning stages of the algorithm, such that they optimize different objective functions, however it is more elegant, as the FEEF does, is that it enables the optimization of the same objective function for both inference and planning, thus casting them as simply different facets of the same underlying process. Moreover, a more serious issue is that this assumption also eliminates the information gain term in active inference -- since the prior and posterior are the same, the divergence between them (which is the information gain), must be zero.

A slightly different approach is taken in a proof in \citep{parr2019computational}, which begins with the KL divergence between two distributions, one encoding beliefs about future states and observations, and the other being the biased generative model. By definition, this KL divergence is always $\geq 0$, which allows us to write.
\begin{align*}
    & \mathbf{D}_{KL} [ p(o_\tau,x_\tau | \pi) \Vert \tilde{p}(o_\tau,x_\tau)] \geq 0 \\
    &= \mathbb{E}_{p(o_\tau | \pi)} \mathbf{D}_{KL} [ p(x_\tau | o_\tau) \Vert \tilde{p}(o_\tau,x_\tau)] - \mathbb{E}_{p(o_\tau|\pi)} [\ln p(o_\tau|\pi) ] \geq 0 \\
    &\Rightarrow - \mathbb{E}_{p(o_\tau | \pi)} \mathbf{D}_{KL} [ p(x_\tau | o_\tau) \Vert \tilde{p}(o_\tau,x_\tau)]  \geq - \mathbb{E}_{p(o_\tau|\pi)} [\ln p(o_\tau|\pi) ] \\
    &\Rightarrow \mathbf{FEF} \geq - \mathbb{E}_{p(o_\tau|\pi)} [\ln p(o_\tau|\pi) ]
\end{align*}

Under the assumption that $p(x | o) \approx Q(x | \pi)$, this becomes:
\begin{align*}
    & - \mathbb{E}_{p(o_\tau | \pi)} \mathbf{D}_{KL} [ p(x_\tau | o_\tau) \Vert \tilde{p}(o_\tau,x_\tau)]  \geq - \mathbb{E}_{p(o_\tau|\pi)} [\ln p(o_\tau|\pi) ] \\
    & \approx \mathbb{E}_{p(o_\tau | \pi)} \mathbf{D}_{KL} [ Q(x_\tau | \pi) \Vert \tilde{p}(o_\tau,x_\tau)]  \geq - \mathbb{E}_{p(o_\tau|\pi)} [\ln p(o_\tau|\pi) ] \\
    & \approx \mathbf{EFE}  \geq -\mathbb{E}_{p(o_\tau|\pi)} [\ln p(o_\tau|\pi) ]
\end{align*}

This proof derives the FEF as a bound on, not the expected model evidence by our definition, but on the entropy of expected observations given a policy. The EFE is then derived from the FEF by assuming that the prior and posterior are the same, which comes with all the drawbacks explained above. This proof is primarily unworkable because of the assumption that the prior and the posterior are identical. While this may be arguable in the continuous time limit, where it is equivalent to the assumption that that $\frac{dQ(x|o)}{dt} \approx 0$, which is when the continuous-time inference has reached an equilibrium, it is definitely not true in discrete time, where although there is a relation between the prior in the current time-step and the posterior in the previous one, it must be mapped through the transition dynamics -- $Q(x_t | \pi) = \mathbb{E}_{Q(x_{t-1} | \pi)}[p(x_t | x_{t-1},\pi)]$.

One can also attempt a related proof by splitting the KL divergence the other way. This gives you:
\begin{align*}
    & \mathbf{D}_{KL} [ p(o_\tau,x_\tau | \pi) \Vert \tilde{p}(o_\tau,x_\tau)] \geq 0 \\
    &= \mathbf{D}_{KL} [ p(o_\tau,x_\tau | \pi) \Vert \tilde{p}(x_\tau| o_\tau)] - \mathbf{E}_{p(o_\tau | \pi)}[\ln \tilde{p}(o_\tau)] \geq 0 \\
    &\Rightarrow - \mathbf{D}_{KL} [ p(o_\tau,x_\tau | \pi) \Vert \tilde{p}(x_\tau| o_\tau)]  \geq - \mathbf{E}_{p(o_\tau | \pi)}[\ln \tilde{p}(o_\tau)] \\
    &\Rightarrow \mathbf{E}_{p(x_\tau | \pi)}[ \ln p(o_\tau | x_\tau)]+ \mathbf{E}_{p(o_\tau | \pi)}\mathbf{D}_{KL} [ p(x_\tau |o_\tau) \Vert \tilde{p}(x_\tau| \pi)]  \geq - \mathbf{E}_{p(o_\tau | \pi)}[\ln \tilde{p}(o_\tau)] \\
    &\Rightarrow \mathbf{FEF} \geq - \mathbf{E}_{p(o_\tau | \pi)}[\ln \tilde{p}(o_\tau)]
\end{align*}

Which is just another way of showing that the FEF is a bound on the expected model evidence.

\section{Related Quantites}
\label{related}

Recently a new free-energy, the generalised free energy (GFE) \citep{parr2019generalised}, has been proposed in the literature as an alternative or an extension to the EFE. The GFE shares some close similarities with the FEEF. Both fundamentally extend the EFE by proposing a unified objective function which is valid for both inferencce at the current time and planning into the future, whereas the EFE can only be used for planning. Moreover, both GFE and FEEF encode future observations as latent unobserved variables, over which posterior beliefs can be formed. Moreover agents maintain prior beliefs over these variables which encode its preferences or desires \footnote{To help make clear the similarity between the GFE and the FEEF, we have defined the veridicial generative model as $Q(o_\tau, x_\tau)$}.

The generalised free energy is defined as 
\begin{align*}
    \mathbf{GFE} = \mathbb{E}_{Q(o_\tau,x_\tau)}[\ln Q(o_\tau) + \ln Q(x_\tau) - \ln \tilde{p}(o_\tau,x_\tau)]
\end{align*}
Whereas the FEEF is defined as
\begin{align*}
    \mathbf{FEEF} = \mathbb{E}_{Q(o_\tau,x_\tau)}[\ln Q(o_\tau,x_\tau) - \ln \tilde{p}(o_\tau,x_\tau)]
\end{align*}

There are two key differences mathematically and intuitively between the GFE and the FEEF. The first is that the GFE maintains a factorised posterior over beliefs and observations, where the posterior beliefs of the two are separated by a mean field approximation and assumed to be separate. By contrast the FEEF maintains a joint approximate belief over both observations and states simultaneously. This joint in the case of the FEEF effectively functions as a veridicial generative model since $Q(o | x) = p(o | x)$ and $Q(x) = \mathbb{E}_{Q(x_{t-1} | \pi)}p(x_t | x_{t-1})$. This is means that posterior beliefs of the future are computed simply by rolling forward the generative model given the beliefs about the current time. 

A second and more important differences lies in the generative models. The GFE assumes that the agent is only equipped with a single generative model with both veridicial and biased components. The preferences of an EFE agent are encoded as a separate factorisable marginal over observations. This means that the generative model of the GFE agent factorises as $\tilde{p}(o,x)_{GFE} \propto p(o | x)p(x)\tilde{p}(o)$. This means that for the GFE the likelihood and the prior are unbiased and there is simply an additional prior preferences term in the free-energy expression. By contrast, the FEEF eschews this unusual factorisation of the generative model and instead presupposes a separate warped generative model for use in the future which is intrinsically biased. The FEEF generative model thus decomposes as $\tilde{p}(o,x)_{FEEF} = \tilde{p}(o|x)\tilde{p}(x)$, which is the standard factorisation of the joint distribution in a generative model, but where both the likelihood and prior distributions are biased towards generating more favourable states of affairs for the agent. This inherent optimism bias then drives action.

A further free-energy proposed in the literature has been the Bethe free-energy and the Bethe approximation \citep{schwobel2018active}. This approach eschews the standard mean field assumption on the approximate posterior in favour of a Bethe approximation from statistical physics \citep{yedidia2001generalized,yedidia2005constructing} which instead represents the approximate posterior as the product of pairwise marginals, thus preserving a constraint of pairwise temporal consistency which the mean-field assumption lacks. Due to this greater representation of temporal constraints (the approximate posteriors at each time-step being no longer assumed to be independent), the Bethe free-energy has the potential to be significantly more accurate than the standard mean-field variational free energy (and is, in fact, exact for factor graphs without cycles such as the standard non-hierarchical POMDP model). In this paper, we focus entirely on the standard mean-field variational free-energy used in the vast majority of active inference publications, and thus the Bethe free-energy is out of scope for this paper. However, exploring the nature of any intrinsic terms which might arise from the Bethe free-energy is an interesting avenue for future work.
Although primarily focused on the Bethe free-energy, \citep{schwobel2018active} also introduced a `predicted free energy' functional. This functional is equivalent to the FEF as we have defined it here, and so has a complexity instead of an information gain term, leading to minimizing the prior-posterior divergence.

Finally, \citep{biehl2018expanding} suggested that if the EFE is not mandated by the free-energy principle, which we have argued for in this paper, then in theory any standard intrinsic measure, such as empowerment, could be used as an objective. We believe that exploring the effect of these other potential loss functions could be a area of great interest for future work.

\end{document}